# Wavelet-based temporal models of human activity for anomaly detection in smart robot-assisted environments[1]

Manuel Fernandez-Carmona [a,*], Sariah Mghames [b] and Nicola Bellotto [b,c]
[a] *Ingeniería de Sistemas Integrados Group, University of Málaga-UMA, Spain*
*E-mail: mfcarmona@uma.es*
[b] *Department of Computer Science, University of Lincoln, United Kingdom*
*E-mails: smghames@lincoln.ac.uk, nbellotto@lincoln.ac.uk*
[c] *Department of Information Engineering, University of Padua, Italy*
*E-mail: nbellotto@dei.unipd.it*

Received 3 April 2023
Accepted 28 August 2023

**Abstract.** Detecting anomalies in patterns of sensor data is important in many practical applications, including domestic activity monitoring for Active Assisted Living (AAL). How to represent and analyse these patterns, however, remains a challenging task, especially when data is relatively scarce and an explicit model is required to be fine-tuned for specific scenarios. This paper, therefore, presents a new approach for temporal modelling of long-term human activities with smart-home sensors, which is used to detect anomalous situations in a robot-assisted environment. The model is based on wavelet transforms and used to forecast smart sensor data, providing a temporal prior to detect unexpected events in human environments. To this end, a new extension of Hybrid Markov Logic Networks has been developed that merges different anomaly indicators, including activities detected by binary sensors, expert logic rules, and wavelet-based temporal models. The latter in particular allows the inference system to discover deviations from long-term activity patterns, which cannot be detected by simpler frequency-based models. Two new publicly available datasets were collected using several smart-sensors to evaluate the approach in office and domestic scenarios. The experimental results demonstrate the effectiveness of the proposed solutions and their successful deployment in complex human environments, showing their potential for future smart-home and robot integrated services.

## 1. Introduction

Modelling temporal series is important to represent sensor data in robotics and many other technical applications, for example to monitor and understand human behaviours with wireless sensor networks in smart environments [34]. The nature of the process, the amount of data required and the extent of the forecasting challenge determine the kind of model finally chosen. Temporal models should be able to capture the frequencies of important event occurrences – e.g. the daily routine of activities performed by an elderly person at home [4]. Methods for frequency analysis (i.e. Fourier transform) can reveal periodic patterns in the sensor data but, if occurring only within specific time intervals, they fail to determine when these periodicities start and end. In this paper, therefore, we propose a new wavelet-based method that is suitable for modelling sparse periodic and/or very short events in sensor data.

[1]The research leading to these results has received funding from the EU H2020 Programme under grant agreement No. 643691, ENRICHME.
[*]Corresponding author. E-mail: mfcarmona@uma.es.

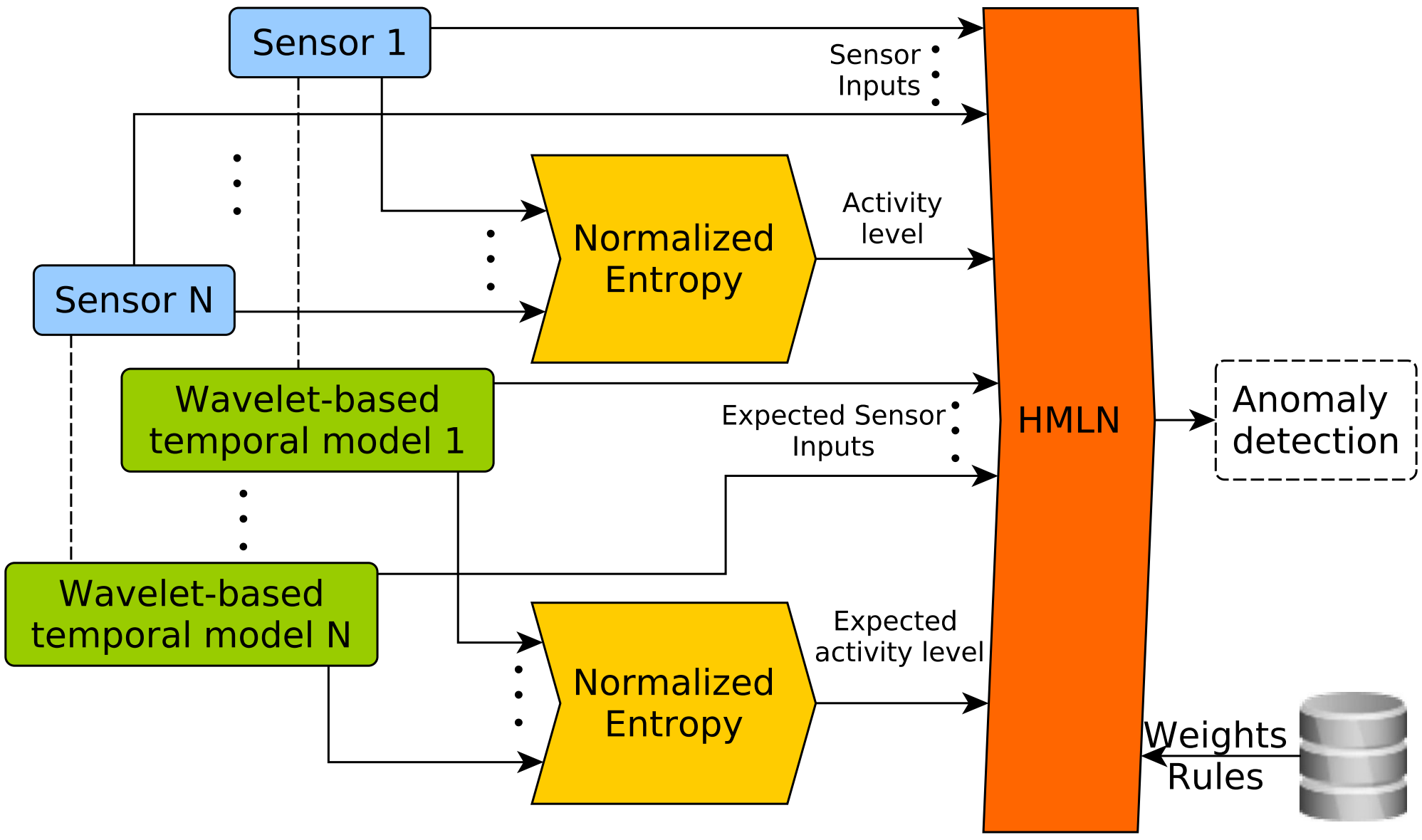

Fig. 1. Wavelet-based anomaly detection system. The expected and actual normalized entropies from wavelet-based temporal models and sensor data, respectively, are compared by a HMLN-based inference module that contains expert rules. In particular, for N sensors (blue boxes), N wavelet-based temporal models (green boxes) are created and used to predict the expected output of each sensor. Real and expected outputs are used to compute, respectively, the real and expected normalized entropies (yellow blocks), which in turn represents real and expected levels of human activity in the environment. These two activity levels, together with the original sensor data, are finally processed by a rule-based hybrid Markov logic network (HMLN, orange block) to detect potential anomalies.

Wavelet analysis has been successfully used for time series classification [23], showing the advantage of simultaneously providing temporal and frequency information of a signal with very little loss of information, and it is therefore more powerful than standard Fourier analysis in capturing and forecasting sensor data for many real-world applications. One such application is Active Assisted Living (AAL) and Assistive Robotics, where good temporal representations of human activities can enable the implementation of many useful well-being services [14].

However, the use of wavelet-based techniques for activity monitoring in AAL scenarios has not been sufficiently investigated in previous literature. To this end, the wavelet-based temporal model proposed in this paper can be used to identify patterns of human activity from the smart sensors of a robot-assisted environment, detecting anomalies that could trigger opportune robot interventions (e.g. if an elderly user spends more time in bed than usual). In this context, anomalies are defined as deviations from typical human activities. The latter have a significant temporal component, which is often periodic, but with occasional variations and possibly very short-term events (e.g. repeatedly opening/closing the fridge in the morning, but only on weekdays). In particular, we adopt the *anomaly* definition by [22], which considers the amount of motion in specific locations as a normalized entropy beyond some given thresholds. Note that the term "motion" is used in a broad sense to include the activation of various binary sensors, such as passive infrared (PIR) motion detectors or contact sensors on doors, cupboards, etc. We also refer to this type of motion in the environment as *activity level*, or intensity.

In this work, we apply our wavelet-based representation of human activities to a new anomaly detection system for AAL (see Fig. 1), complementing and extending our previous assistive robotic system [14]. The system predicts sensor data and detect anomalies when the expected intensity of human activities differs from the real one. In particular, given a set of smart-home sensors (i.e. motion detectors, temperature, light and contact sensors), we build accurate temporal models to represent and forecast their expected output. Then, using an entropy-based method [22], we estimate the current and expected levels of human activity. These two levels are finally compared by a new inference system, based on a Hybrid Markov Logic Network (HMLN) [54], to detect and integrate "statistical" anomalies with user-defined expert rules.

The paper includes three main contributions:

- First, we describe a new technique for temporal modelling of (long-term) human activities based on wavelet transforms. Among its possible applications, this wavelet-based temporal model enables the forecasting of smart-sensor signals for the detection of potential anomalies, i.e. human activities that deviate significantly from the norm. A software implementation of this temporal modelling tool is made publicly available.
- Second, we propose an automatic system for anomaly detection that uses a HMLN to combine three sources of information about human activities, namely i) actual entropy level from smart-home sensors, ii) expected entropy from wavelet-based temporal models, and iii) expert knowledge in the form of logic rules.
- Finally, we present extensive experimental results based on two large datasets, one previously recorded in an office environment [22] and a new one from a real elderly home, which we also made publicly available. These datasets were recorded in MongoDB format [9] for easy access and re-usability by the scientific community.

The remainder of the paper is organized as follows. Section 2 reviews state-of-the-art methods for temporal modelling and anomaly detection with smart-home sensors, including relevant public datasets. Section 3 briefly introduces the wavelet transform and describes the respective temporal models of sensor data. Section 4 describes the design of the HMLN-based inference systems and its expert rules to analyse and detect anomalies in human activities. Section 5 illustrates the architecture and practical implementation of the anomaly detection system. Section 6 presents datasets and experiments to validate the effectiveness of the temporal models and the anomaly detection in an office and AAL scenarios. Finally, Section 7 discusses advantages and disadvantages of the proposed approach, suggesting directions for future work in this area.

## 2. Related work

Temporal modelling is widely used to detect regular patterns in data. From time series analysis, a relevant tool is the *autoregressive integrated moving average model* (ARIMA) used, for example, for applications as varied as stochastic wind power modelling [8] and, more recently, for the prediction of human behaviours in smart-home settings [53].

The main problem with these models, though, is that they are only suitable for relatively short temporal windows or known temporal trends [58]. Other non-linear techniques, such as Gaussian Processes, could theoretically achieve the full reconstruction of signals from mixture models. For example, Ghassemi & Deisenroth [25] use periodic Gaussian Processes for long-term forecasting. Similarly, recent work by [27] has used a non-parametric prognostic framework for individualized event prediction based on Poisson processes with a multivariate Gaussian convolution method. These models are typically robust against model instabilities, but they require heavy computational processes.

A technique called FreMEn (Frequency Map Enhancement) has also been proposed for spatio-temporal representations of robot environments in long-term scenarios [30]. It uses Fourier analysis to extract periodicities in sensor data, in combination with a Bernoulli distribution or Poisson processes [29] to represent binary information states. FreMEn is a simple yet effective modelling tool, but it is not suitable to describe sparse or very short events.

Wavelet-based methods have been used for temporal analysis in many different fields such as drought or price forecasting [11], passenger flow prediction [49], human gait detection [2], motion and texture classification [16]. Since wavelets contain both frequency and time domain information, they are particularly suitable to represent sparse non-stationary signals. Indeed, wavelet transforms have been successfully used to model complex data patterns in sensor networks and wearable systems for human activity recognition [45,55]. Their application to sensor modelling and forecasting for activity monitoring in long-term studies, however, are still underinvestigated. Given their properties, it is particularly interesting to evaluate whether they can reliably characterise short- and long-term human activities in smart-home AAL settings.

Soulas et al. [47] proposed an Extended Episode Discovery model that defines habits in terms of length, frequency and periodicity for offline processing. In [7], the authors compare three sequential activity models – Hidden

Markov Model (HMM), Conditional Random Fields (CRF) and sequential Markov Logic Network (MLN) – where feature vectors were generated during fixed-time windows for on-line processing. Although potentially useful, these sequential activity models were not used for anomaly detection in the aforementioned works.

Typically, anomaly detection systems are designed for the specific sensor(s) used. Depending on the input data, approaches may vary greatly. Wearable activity trackers like the ones used for human activity recognition [37,45], for example, provide continuous motion and pose information without requiring any additional preprocessing.

Markov Logic Networks (MLNs) are both a modelling [32] and inference [28] tool, often used for their flexibility to define rich models. They are able to perform inference using imprecise or incomplete inputs, useful to deal with sensor faults and network errors. In addition, they can blend both sensor data and expert logic rules within a probabilistic framework for robust inference in real-time applications [50]. Compared to SVM and HMM-based systems, the advantage of using MLNs for anomaly detection is that they require a smaller amount of sensor data to build their models and that they better handle uncertain information [24]. SVM have been successfully combined with deep learning techniques for anomaly detection and achieved promising results in high dimensional problems [18], but without exploiting the available temporal information.

An unsupervised anomaly detection for multivariate time series was proposed in [1] based on an adversarially-trained auto-encoder that outperformed previous unsupervised methods, such as Isolation Forests, Autoencoders, LSTM-VAE, DAGMM [62], and OmniAnomaly [48]. A variational representation of an encoder-decoder based LSTM (VLSTM) was proposed in [61] for a supervised anomaly detection in industrial big data, efficiently coping with typical issues of imbalanced high-dimensionality. However, in addition to the significant effort required for labeling, VLSTM was not benchmarked against other frameworks for anomaly detection. The authors in [57] proposed an LSTM-Autoencoder (LSTM-AE) for unsupervised anomaly detection in indoor environments, evaluated only on a single univariate dataset. Another unsupervised attention-based convolutional LSTM-based autoencoder with dynamic thresholding (ACLAE-DT) was proposed by [51] for anomaly detection and diagnosis in multivariate time series applied to smart manufacturing. ACLAE-DT showed better performance compared to other baselines, such as SVM and ARIMA (Auto-Regressive Integrated Moving Average), but it required a large amount of training data, further augmented by a pre-processing phase to capture the intercorrelation between sensor pairs. Although these unsupervised methods achieve good performance, they typically require large amounts of data in order to generalise well, which are not always available. Moreover, it remains unclear how to fine tune and incorporate additional expert knowledge into such models to fit a particular applications.

A different non-deep approach for fast, online and accurate anomaly detection was proposed in [56] leveraging an energy-based extreme learning machine for online detection of anomalies in traffic signals. This framework outperformed SVM, artificial neural networks, K-NN, and Random Forest on a DDoS (distributed denial-of-service attacks) dataset. Other works [10,39,43] addressed the explainability, and therefore causes, of detected anomalies. In particular, the approach proposed by [10] achieved the best prognostic performance among several baseline methods for anomaly detection. One of the most recent works in unsupervised anomaly detection for time series data was presented by [41] comparing six deep and traditional methods. The results show that in many cases classical machine learning methods can outperform (in terms of runtime and accuracy) many deep-learning solutions across a wide range of anomaly types. Motivated by the results of this latter study, our paper propose a non-deep architecture based on HMLN that combines wavelet-based temporal models and expert rules, mixing for the first time discrete and continuous predicates, to infer potential anomalies. These expert rules allow also to overcome the lack of data otherwise required to train deep learning-based methods.

Obviously, public datasets with labelled sensor data are important to test and compare different algorithms. Unfortunately though, there are not many such datasets available in the context of smart-sensor networks, especially for AAL applications. The dataset by Tim van Kasteren [2] [52] offers a collection of compressed Matlab files with several recordings of binary sensors (e.g. open/closed doors; pressure mats; motion detectors). The Center for Advanced Studies in Adaptive Systems (CASAS) also provides an extensive collection of datasets [3] [12] for activity recognition, in which every entry has a different format, usually a compressed text or binary file. Both these datasets, however, contain non-standard, plain text, or binary files which are difficult to handle by other researchers, especially if the dataset size is large. They also lack a standardized format and access mechanisms, suitable for systematic processing of big data. To our knowledge, there are no smart-home datasets based on such standardised and easily manageable formats. Our new dataset, instead, was created by storing raw data in a MongoDB database. This approach provides an accessible, platform- and application-independent format readily available for further research in our application area and beyond.

[2] http://ailab.wsu.edu/mavhome/research.html

[3] http://casas.wsu.edu/datasets/

## 3. Wavelet-based temporal forecasting

In this section we present a novel approach to forecast sensor data for human activity monitoring using a wavelet-based temporal model. We start with a brief description of the discrete wavelet transform algorithm, and then we explain how to tune and use this algorithm for building our temporal model of the sensor data.

Wavelets provide an alternative representation that overcomes the limitations of Fourier analysis. They decompose signals into individual components, which maintain both frequency and time information. Also, they can effectively represent and provide localized information about discontinuities. These advantages (i.e. time-frequency and discontinuity representations) are very important to handle the non-periodic and often "spiky" nature of real-world sensor data, especially in the context of activity monitoring.

### 3.1. Discrete wavelet transform

A discrete wavelet transform (DWT) is a sampled wavelet transform applicable to digital signals. It uses two sets of weighted and displaced reference signals (wavelets $\varphi_{j,k}$ and $\psi_{j,k}[n]$) to represent a discrete time signal $x[n]$ using the following decomposition:

$$x[n] = \sum_{k=0}^{\frac{N}{2}-1} c_{Q,k}\varphi_{Q,k}[n] + \sum_{j=1}^{Q}\sum_{k=0}^{\frac{N}{2}-1} d_{j,k}\psi_{j,k}[n] \tag{1}$$

This is, using the averaging $c_{Q,k}$ and detail coefficients $d_{j,k}$ we can fully describe the signal $x[n]$ using an smaller amount of information.

### 3.2. Sensor data modelling and forecasting

After introducing the wavelet transform and its parameters, we can use them to model smart-home sensors and to forecast their data. Our model is an efficient representation of a generic temporal signal, similar to some compression techniques commonly used in image processing [20].

Let us consider the signal $x[n]$ generated by a smart-home sensor over time, which can be either binary (e.g. presence detectors, door contacts, switches) or continuous (temperature, energy consumption, light). The sampling frequency of the sensor data is $f_s$, that is, the sensor provides a reading every $1/f_s$ seconds. We want to build one wavelet model for each sensor. In order to do that, our training model signal is transformed into the wavelet domain using a 1-level DWT decomposition. Since the input data is relatively sparse (i.e. mostly containing localized activation peaks), a higher decomposition level would not bring any particular advantage to the resulting wavelet transform. We then threshold the wavelet coefficients and keep a significantly smaller number of them, while maintaining a low Root Mean Square Error (RMSE). We can finally reconstruct the signal using this small subset of coefficients and the inverse wavelet transform (IDWT).

Our wavelet-based model $\mathcal{M}$ is therefore described by this subset $\hat{C}$ of coefficients, a mother wavelet $\varphi$, the decomposition level $Q$, a coefficient threshold $\tau$, the number of samples $N$, the sampling frequency $f_s$, and the time reference $t_0$:

$$\mathcal{M} = \{\hat{C}, \varphi, Q, \tau, N, f_s, t_0\} \tag{2}$$

Once this model is available, it is possible to represent the sensor output at a future time instant $t_f$. The model in (2) assumes that the sensor output has periodicity $N$ starting from time $t_0$. The index $n_i$ of the sensor data sample at time $t_f$ is therefore given by the following equation:

$$n_i = (t_f - t_0) * f_s \mod N \tag{3}$$

and the actual sensor data sample can then be obtained from the reconstructed signal $\hat{x}[n]$ as follows:

$$\hat{x}[n_i] = IDWT(\hat{C}, \varphi, \psi)[n_i] \tag{4}$$

In practice, we will use the set of coefficients $\hat{C}$ above a certain threshold $\tau$ that still allows a lossless reconstruction of the signal [38]. All the remaining coefficients, below the selected threshold, will be removed from our sensor data model. We therefore chose the lowest threshold value ($\tau = 0.54$) for all the wavelet sensor models that allows full reconstruction of the signals in the training dataset.

## 4. Anomaly detection

Markov Logic Networks can be used to combine different sources of information for probabilistic inference. In this paper, we use both smart-home motion sensors and their wavelet-based models to analyse the difference between *actual* and *expected* entropy, respectively, of the environment. The first one represents the current activity level, whereas the second one represents the most likely one. These entropy values, together with direct sensor inputs and expert rules, provide the necessary information for our MLN to detect anomalous situations, as shown also in Fig. 1.

### 4.1. Hybrid Markov logic networks

MLNs combine both probabilistic and logical reasoning [42]. Briefly, a MLN consists of a set of weighted first-order logic formulas or clauses. The latter include the following elements:

- *constants*, which are possible objects in the domain of interest;
- *variables*, describing a set of objects in that domain;
- *functions*, mapping relations between different objects;
- *predicates*, defining logical attributes or relationships over the domain's elements, which can be combined into more complex formulas using logical connectors.

Functions, variables and constants are called *terms*. If they do not contain variables, they are *ground terms*. A predicate that contains only ground terms is a *ground predicate*. When a logical value is assigned to all grounded predicates in a network, we have a *possible world*.

Using evidences, MLNs can produce Markov networks that describe the probability of all possible combinations of grounded clauses. We can then perform inference on these Markov networks, usually by using approximate methods such as MC-SAT [54]. Besides discrete evidence value, it is also possible to consider continuous ones using an extension called Hybrid Markov Logic Network (HMLN) [54]. Thanks to the latter, we can thus consider predicates based on continuous variables that contain our entropy values of the activity levels.

### 4.2. Wavelet model as prior for HMLN

According to caregivers, some of the most concerning behaviours of people with cognitive impairments are related to memory losses [5], wandering and repetitive actions [15]. Memory losses may be linked with problematic behaviours, for example if user leaves some doors open. If user leaves main door open it will likely affect the house temperature and be a safety concern. Similarly, if the fridge door is left open, it will increase the energy consumption, break the fridge itself and spoil the food. Wandering may be another concerning behaviour, specially if happens during resting hours where there are no caregivers around and accidents are more likely to happen. But overall, wandering during resting hours keeps patients from resting. Our system is able to detect those situations, considered as "inappropriate behaviours" using logical rules.

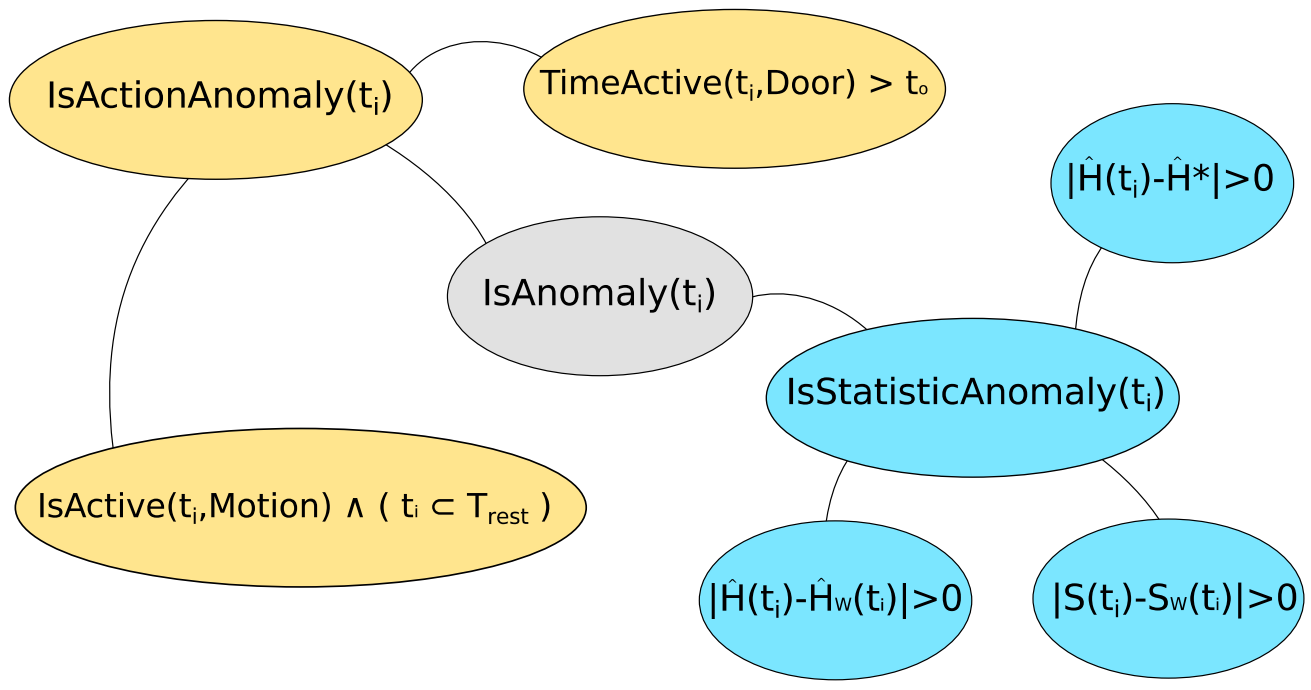

Fig. 2. Grounded Markov network from the predicates in (5), (6) and (7). The blue nodes capture statistical differences between current and expected values from sensors and activity levels. The yellow nodes instead implement ad-hoc expert rules.

Repetitive actions are also a source of concern, as they are related to dementia. Changes on the intensity of the activity levels or usual times can be indicators of a cognitive decline. For example, if users change they activity pattern spending more time in the kitchen than usual or at unusual times -for them-, it may related to dementia episodes. The wavelet-based sensor data model defined in Section 3.2 can be used to predict the expected output of a particular sensor based on historical data and detect "Statistical anomalies" using logical rules as well.

These predictions $S_W(t_i)$ at time $t_i$ are the expected values, either binary or continuous, that can be compared with the current sensor values $S(t_i)$. From the expected output of all the sensors, it is also possible to compute the normalized entropy $\hat{H}_W(t_i)$ that represents the *expected activity level* for the whole environment (see Fig. 1). The entropy $\hat{H}(t_i)$ from all the real sensors represents instead the current activity level. These two activity levels, current and expected, are compared by the following HMLN to determine whether an anomalous situation is occurring.

We define three clauses to combine our sources of information: one to check whether the current sensor value $S(t_i)$ is above its expected $S_W(t_i)$; a second one to test if the current $\hat{H}(t_i)$ is above a certain threshold $\hat{H}^*$ (defined by a percentile of the average entropy for all $t < t_i$); and another to compare $\hat{H}(t_i)$ to the expected $\hat{H}_W(t_i)$ (i.e. in case there are some "surprising" events). The occurrence of any of these conditions indicates a potentially anomalous situation at time $t_i$, captured by the predicate `IsStatisticAnomaly`:

$$
\begin{aligned}
&S(t_i) - S_W(t_i) \geqslant 0 \Rightarrow \texttt{IsStatisticAnomaly}(t_i) \\
&\hat{H}(t_i) - \hat{H}^* \geqslant 0 \Rightarrow \texttt{IsStatisticAnomaly}(t_i) \\
&\hat{H}(t_i) - \hat{H}_W(t_i) \geqslant 0 \Rightarrow \texttt{IsStatisticAnomaly}(t_i)
\end{aligned}
\tag{5}
$$

In our system, $\hat{H}^*$ is the 90% percentile of the average entropy from the real sensors. This threshold is generally accepted as a statistically meaningful indicator of anomaly, and it has been used to detect outliers in financial time series using wavelets [26] or among many response time outlier detectors in web surveys [35]. The predicate in (5) and its clauses are represented by the blue connected nodes in Fig. 2, which shows the graph of a grounded HMLN at time $t_i$.

An advantage of MLNs is that they can combine different logical rules. This allows us to combine these "Statistical anomalies" rules with new ones describing "inappropriate behaviours". For AAL applications, such rules could be provided by clinicians or professional carers and adapted to the specific person being monitored. In our system these can be monitored by means of motion detector and contact sensors on doors and appliances. Their outputs determine the state of the predicate `IsActionAnomaly`, which is implemented in our HMLN as follows (see also yellow nodes in Fig. 2):

$$
\begin{aligned}
&\texttt{TimeActive}(t_i, \textit{Door}) > t_0 \Rightarrow \texttt{IsActionAnomaly}(t_i) \\
&\texttt{IsActive}(t_i, \textit{Motion}) \wedge (t_i \subset T_{rest}) \Rightarrow \texttt{IsActionAnomaly}(t_i)
\end{aligned}
\tag{6}
$$

where *Door* is a contact sensor, *Motion* is a motion detector, $t_0$ is the minimum time of a door left open to be considered an anomaly, and $T_{rest}$ is the resting time interval suggested by a human expert (e.g. 11:00 P.M. to 7:00 A.M.). The first rule describes a door that is left open, and the second rule describes an unusual wandering behaviour during resting hours.

The two types of anomaly are finally combined by the following `IsAnomaly` predicate (central node in Fig. 2):

$$\mathtt{IsStatisticAnomaly}(t_i) \vee \mathtt{IsActionAnomaly}(t_i)$$
$$\Rightarrow \mathtt{IsAnomaly}(t_i) \quad (7)$$

In conclusion, we have two groups of expert rules: the first group is related to sensors and their outputs – `StatisticAnomaly` – focusing on deviations from routines and patterns; the second group represents "inappropiate behaviours" – `ActionAnomaly`. In Section 6.4 we will evaluate the anomaly detection with and without the contribution of the expert rules in (4.2) to better understand the contribution of the wavelet- and entropy-based methods.

## 5. System implementation

The solutions described in the previous sections have been implemented in ENRICHME, [4] a research project integrating ambient intelligence and robotics to provide AAL services for elderly people with mild cognitive impairments [4,14]. The ENRICHME system monitors the activity of these people at home, exchanging information between a network of smart-home sensors, a mobile robot and an auxiliary Ambient Intelligence Server (AIS) (see Fig. 3). The latter consists of an embedded PC, located at home, which acts as a multiprotocol gateway, collecting and forwarding the information shared wirelessly between robot and smart-home sensors for monitoring human motion, doors/cupboards use, and energy consumption. The sensor network is based on the Z-Wave communication protocol and uses the OpenHAB middleware, [5] which supports a wide range of different smart-home technologies with a uniform interface, decoupling sensor information from specific smart-home protocols and manufacturers [46].

The embedded PC for data recording and processing is an Intel NUC i7-5557U CPU @ 3.10 GHz with 8 GB of RAM, running Linux OS Ubuntu 14.04 64 bits. The smart-home sensors are commercial Z-Wave wireless devices produced by the Fibar Group. [6] These sensors are small, easily deployable, widely available and have a long battery life.

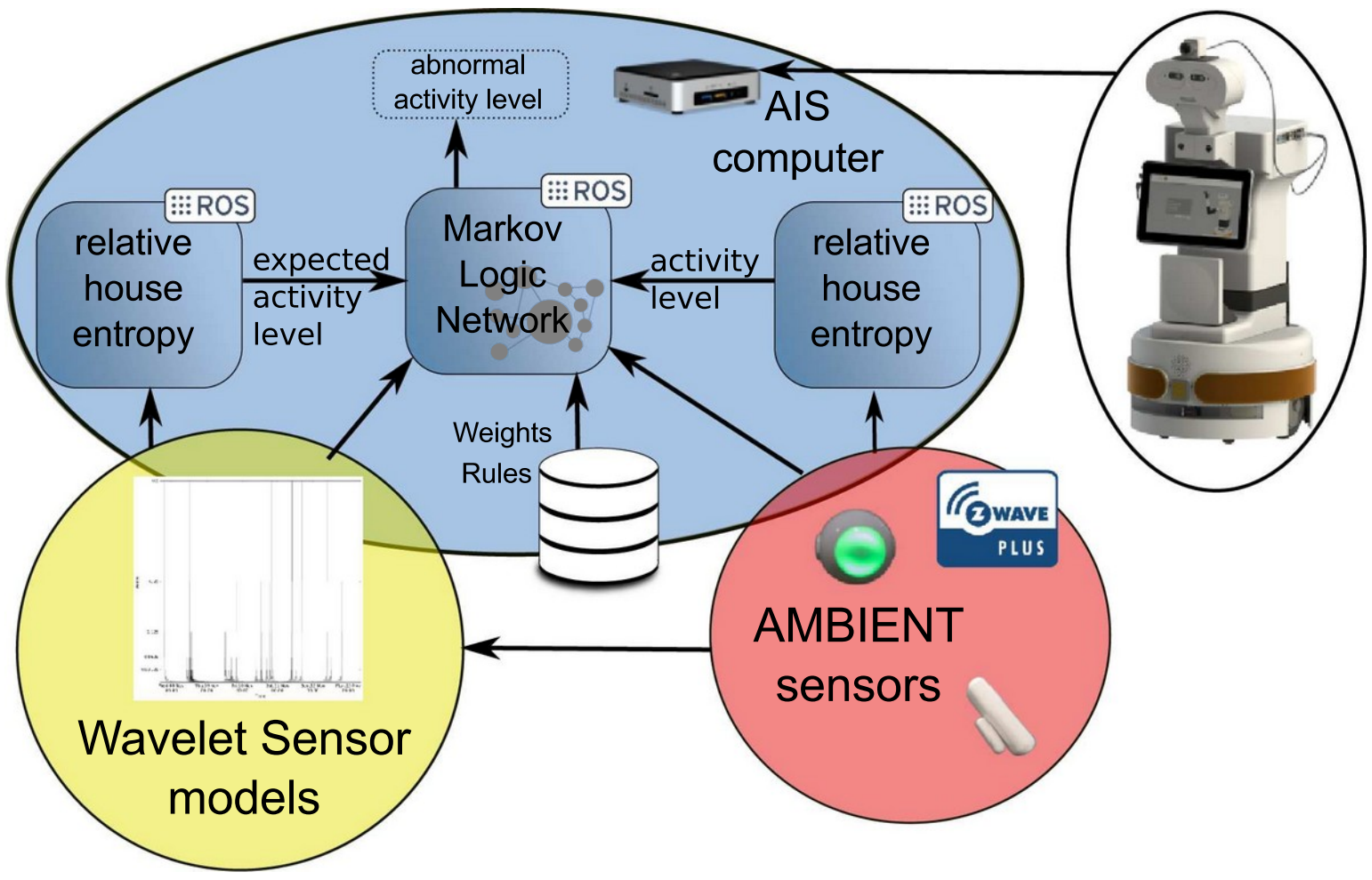

Fig. 3. Smart-home sensors integration in ENRICHME.

[4] https://cordis.europa.eu/project/id/643691
[5] http://www.openhab.org
[6] http://www.fibaro.com

The anomaly detection system is implemented as a Robot Operating System[7] (ROS) module making use of efficient MLN libraries for online inference [21]. ROS provides a common framework for information exchange between AIS and robot, so that the latter can easily access the results of the HLMN inference engine. The HMLN can be queried using evidence provided by any ROS source, including the actual and expected house entropies obtained from the sensors and the wavelet models, respectively. These evidences are gathered using standard ROS service clients, which standarizes the exchange format and guarantees the overall process. The output of the inference process is also available to any other node on the ROS network, for example to trigger a specific robot behaviour or alert a remote telecare system. It is published as a *topic*, meaning that is a one-way information source available for any ROS module. Note that our wavelet-based sensor modelling system is available as a ROS action server. [8] This software allows the creation, management and querying of multiple models. Online reasoning its triggered periodically, comparing sensor output with wavelet models and testing expert rules using HMLN networks. Our statistical rules are able to detect unusual data patterns coming from the sensors, e.g. a light level that is not changing as expected at a specific query time.

The inference service can be also triggered under demand by other ROS modules, for example giving the assistive robot the possibility to evaluate anomalies at will or trigger evaluations not only periodically but whenever particular criterion is met. The proposed system could be potentially used as an *"space-behavior-health"* just by identifying the key spatial areas where anomaly detection should be evaluated.

## 6. Experiments

The performance of our proposed solutions were evaluated using real data recorded from different scenarios. In this sections, we will first describe two different datasets: one already presented in [22] and one newly recorded. Then, we will use them to evaluate the forecasting capabilities of our wavelet sensor model compared to another similar tool in the literature. This is, we will implement wavelet models for all the sensors on each dataset. Based on these wavelet models, we will calculate the expected entropy levels of the testing environments and finally demonstrate their use as priors for anomaly detection. Wavelet models are the base of our anomaly detection system, but HMLNs allow to blend them with expert rules derived from caregivers. So, in this section we will also study the impact of the expert rules described in Section 4.2 in our anomaly detection system. The implemented rules (*'door left open'* and *'activity during resting hours'*) are some of the most common on the literature. Finally, the performance of the anomaly detection system was compared also against some state-of-the-art unsupervised anomaly detection methods.

### *6.1. Sensor datasets*

All the datasets were recorded using MongoDB, an open-source cross-platform document-oriented database. MongoDB is a NoSQL database, using JSON-like documents with schemas. Compared to traditional log and spreadsheet files, this storage approach offers better data management and manipulation, which is particularly important for long-term datasets like ours. MongoDB can work very efficiently with time series and also provides efficient and flexible querying methods, so we can easily retrieve any data interval, sensor set, or even combine data from other sources [40]. Our datasets are recorded from commercial devices managed by OpenHAB, one of the most popular open-source software platforms for home automation, which ensures the accuracy, consistency and completeness of the stored data. The sensor datasets presented here can be seamlessly augmented with and compared to future recordings from a large range of different domotic sensor standards.

The first dataset was collected for 111 days in an office environment (L-CAS dataset [22]) including: a lounge with sofas and a coffee table; a kitchenette with various appliances and cupboards for storing and preparing food; an entrance and a workshop area. This dataset contains data from ten different physical devices, which provided six different types of sensor data readings: humidity, temperature, light, energy consumption, motion, and binary contact (for door activation). The sensors were located in five different locations, and their data recorded approximately every 30 seconds, generating more than 400,000 data entries in the final dataset.

[7] http://www.ros.org
[8] https://github.com/LCAS/wtfacts

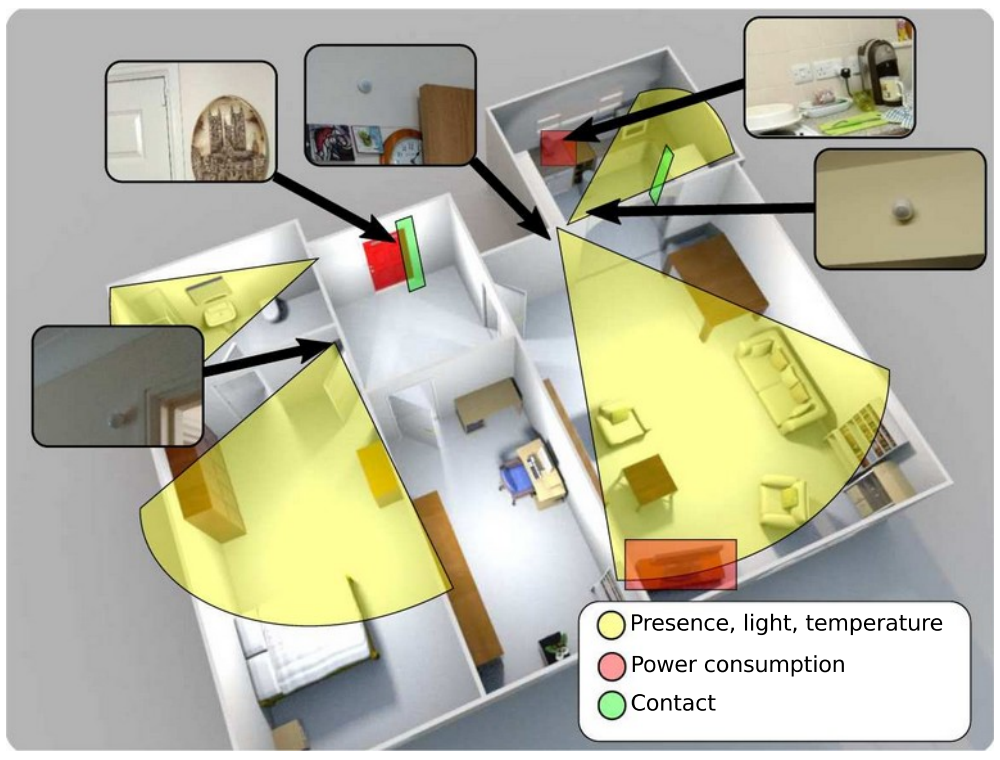

(a) LACE apartment's layout.

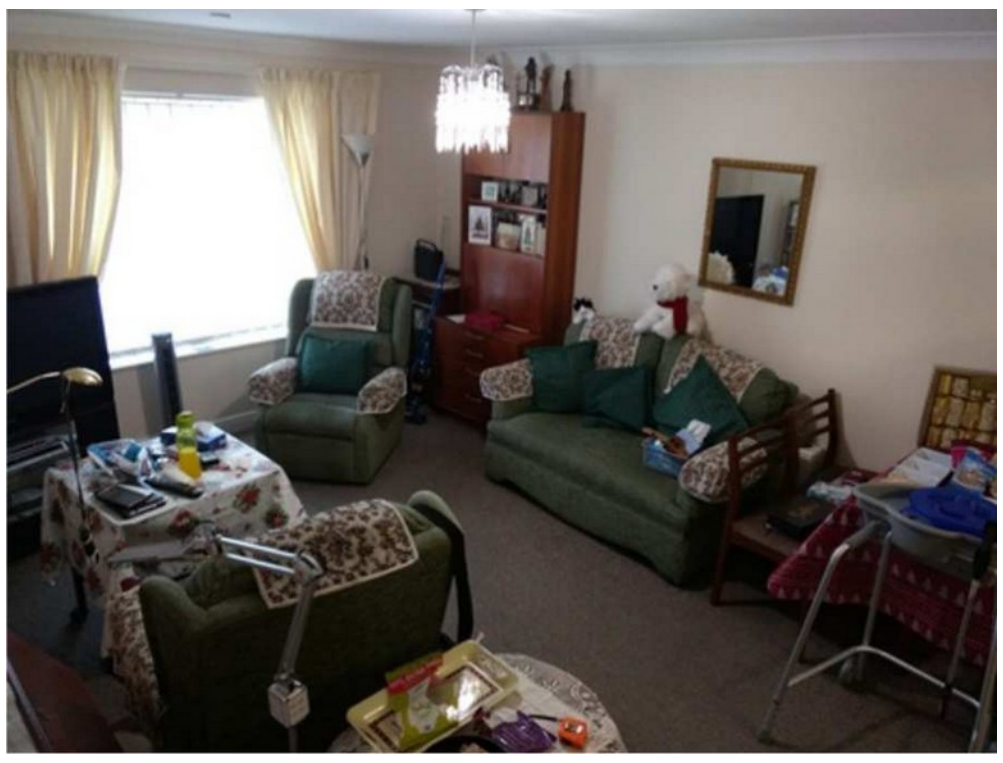

(b) Living room of the LACE apartment.

Fig. 4. ENRICHME dataset environment.

More than ten people were working in the L-CAS premises during the recording. The sensors were mostly concentrated in places where a rich set of activities were typically performed (entering, exiting, eating, drinking, resting, etc.). The dataset is split in two parts: the first one, used for training, includes sensor data continuously recorded for three months and a half; the second one includes one week of data used for testing.

The new dataset was recorded in the apartment of an elderly couple within the residential facilities of LACE Housing[9] as part of the ENRICHME project. It contains one month (31 days) of sensor data with five types of readings (temperature, light, energy consumption, motion and door activation), corresponding to approximately 33,000 entries in total. The sensors covered most of the apartment area, recording data from the entrance, the kitchen, the living room, the main bedroom and the bathroom. Figure 4 illustrates the approximate sensors' position and area coverage. The first three weeks of the dataset were used for training, while the last week for testing.

Table 1 summarizes the duration, locations, sensors, and other general characteristics of the recorded datasets. For comparison, the table includes also the main characteristics of other two datasets, Kasteren [52] and CASAS [12], from the related work (Section 2).

Note that, to evaluate our wavelet-based models and anomaly detection (Section 6.2 and Section 6.4), both continuous and binary sensors were used. However, only the latter were necessary for our entropy-based activity representation (Section 6.3).

### 6.2. *Model training and forecasting*

We divided our datasets (see Table 1) into two folds: one for training and one for testing the prediction. We trained our models using the first part of the datasets. Once the models were ready, we evaluated them by comparing their predicted outputs with the real data from the testing part of the datasets. In the L-CAS dataset, we used the first three months (104 days) of sensor data for training and then one week for testing. The ENRICHME dataset had a smaller number of entries, so we used three weeks for training and 10 days for testing.

For each sensor in a dataset, we took the testing dataset timestamps and forecasted the expected sensor values at those instants using the trained models. For the binary sensors, if both testing data and model forecasting values were 'true', the prediction was considered as a 'True Positive'. Similarly, if both testing data and prediction were 'false', it was a 'True Negative'. Missed activations ('true' in the dataset but 'false' in the prediction) were marked as 'False Negative', and predicted activations that were wrong were considered 'False Positive'. With those four parameters, we computed the statistics summarized in Tables 2 and 3. For the continuous sensors, we computed the RMSE between testing data and forecasted values, as presented in Tables 4 and 5.

[9] http://lacehousing.org

Table 1

Dataset entries summary. In addition to the L-CAS and ENRICHME datasets, two other datasets from the literature are included for comparison

| Dataset | | Sensors | Locations | Num. entries | Data types | People in dataset | Duration (days) | Storage format |
|---|---|---|---|---|---|---|---|---|
| **L-CAS**[*] | | Motion, Binary Contact, Humidity, Light, Energy Consumption, Temperature | Entrance, Fridge, Kitchen, Lounge, Workshop | 492,441 | Binary, Integer, Float | 12 | 104<br>7 | Mongo |
| **ENRICHME**[**] | | Motion, Binary Contact, Light, Energy Consumption, Temperature | Entrance, Fridge, Kitchen, Bathroom, Bedroom, Livingroom, TV | 33,838 | Binary, Integer, Float | 2 | 31 | Mongo |
| Kasteren | House A | Binary Contact, Motion, Toilet Flush, Appliance Activation | 3 rooms | N/A | Binary | N/A | 25 | Plain text |
| | House B | | 2 rooms | | | | 15 | |
| | House C | | 6 rooms | | | | 20 | |
| CASAS | Aruba | Motion, Binary Contact, Temperature | Kitchen, 2 x Bathrooms, Office, Livingroom, 2 x Bedrooms | | Binary, Integer, Float | 1 | 180 | Plain text |
| | HH** | Battery Levels, Binary Contact, Light Switches, Light, Motion, Temperature | N/A | | Binary, Integer | 1 or 2 | up to 180 | |
| | Kyoto DailyLife | Motion, Appliance activation | Livingroom, Kitchen, Bedroom | | Binary | 1 or 2 | up to 6 | |

[*]https://lcas.lincoln.ac.uk/wp/research/data-sets-software/l-cas-domotic-sensors-dataset/
[**]https://lcas.lincoln.ac.uk/wp/lace-house-domotic-sensors-dataset/

Table 2

Comparison between predictions from Wavelet (W), FreMEn (F), ARIMA (A) and CNN-based (C) models in the L-CAS dataset

| Sensor Location | | Presence detectors: Entry | Kitchen | Lounge | Workshop | Sensor Average |
|---|---|---|---|---|---|---|
| Precision (%) | W | **63.1** | **65.4** | **61.1** | **53.6** | **60.8** |
| | F | 48.4 | 50.3 | 44.9 | 37.9 | 45.4 |
| | A | 54.2 | 53.8 | 50.6 | 45.7 | 51.1 |
| | C | 55.0 | 56.9 | 53.1 | 46.3 | 52.8 |
| Recall (%) | W | 57.1 | **69.8** | **62.4** | 51.5 | **60.2** |
| | F | **70.8** | 66.5 | 42.1 | 47.7 | 56.8 |
| | A | 63.4 | 63.8 | 49.5 | 45.9 | 55.6 |
| | C | 63.5 | 68.0 | 53.8 | **54.5** | 60.0 |
| Accuracy (%) | W | **88.3** | **80.2** | **87.4** | **97.6** | **88.4** |
| | F | 59.8 | 58.1 | 70.5 | 72.1 | 65.1 |
| | A | 73.9 | 68.1 | 75.7 | 80.0 | 74.4 |
| | C | 73.6 | 68.6 | 79.5 | 89.0 | 77.7 |
| F1 score (%) | W | **60.0** | **67.5** | **61.7** | **52.5** | **60.4** |
| | F | 57.5 | 57.3 | 43.5 | 42.2 | 50.1 |
| | A | 58.4 | 58.4 | 50.1 | 45.8 | 53.2 |
| | C | 59.0 | 62.0 | 53.5 | 50.1 | 56.1 |

In order to evaluate the prediction quality of our wavelet sensor model, we compared it to other three different forecasting tools, considering frequency-, statistics-, and neural network-based approaches. The first one, called Frequency Map Enhancement (FreMEn) [30], was originally developed for robotics applications but then applied also to smart-home sensors [13]. FreMEn is a method that allows to model periodic changes of the environment using Fourier-based spectral analysis. It considers the probability of the environment's state to be a function of time, represented by a (compressed) combination of harmonic components. The second forecasting tool is based on autoregressive integrated moving average models (ARIMA) [53]. These statistical models combine a regressive part to capture the relation between current and past observations, a moving average component that includes lagged forecasting errors into the prediction, and an integral part to ensure that series are stationary. Finally, the third one is

Table 3

Comparison between predictions from Wavelet (W), FreMEn (F), ARIMA (A) and CNN-based (C) models in the ENRICHME dataset

| Sensor Location | | Presence detectors | | | | Door sensors | | Sensor Average |
|---|---|---|---|---|---|---|---|---|
| | | Bathr. | Bedr. | Kitch. | Living | Entrance | Fridge | |
| Pre. (%) | W | 94.0 | 93.9 | 89.8 | 87.3 | **99.6** | **99.6** | 94.0 |
| | F | **95.8** | **95.6** | 92.8 | 89.5 | 99.7 | **99.6** | **95.5** |
| | A | 92.1 | 93.6 | 89.3 | 86.1 | 98.1 | 95.5 | 92.5 |
| | C | 94.2 | 94.0 | **92.9** | **91.3** | 99.1 | 99.4 | 95.1 |
| Rec. (%) | W | **93.3** | **93.8** | **90.1** | **87.3** | **99.6** | **99.4** | **93.9** |
| | F | 90.8 | 84.2 | 66.5 | 59.0 | 81.3 | 83.1 | 77.5 |
| | A | 87.8 | 84.8 | 75.8 | 71.5 | 85.6 | 89.6 | 82.5 |
| | C | 91.7 | 88.2 | 79.6 | 74.4 | 90.3 | 90.5 | 85.8 |
| Acc. (%) | W | **88.1** | **88.6** | **82.1** | **77.9** | **99.3** | **99.1** | **89.2** |
| | F | 87.7 | 81.5 | 65.5 | 58.4 | 81.1 | 82.9 | 76.2 |
| | A | 83.2 | 80.9 | 71.8 | 64.8 | 87.8 | 88.9 | 79.5 |
| | C | 87.4 | 85.0 | 75.7 | 70.9 | 89.3 | 90.7 | 83.2 |
| F1 (%) | W | **93.7** | **93.9** | **90.0** | **87.3** | **99.6** | **99.5** | **94.0** |
| | F | 93.2 | 89.5 | 77.5 | 71.1 | 89.6 | 90.6 | 85.3 |
| | A | 89.9 | 89.0 | 82.0 | 78.1 | 91.4 | 92.4 | 87.1 |
| | C | 92.9 | 91.0 | 85.7 | 82.0 | 94.5 | 94.7 | 90.1 |

Table 4

RMSE (%) of predictions for continuous sensors using Wavelet (W), FreMEn (F), ARIMA (A) and CNN-based (C) models in the L-CAS dataset

| | | Entry | Kitchen | Lounge | Workshop | Average |
|---|---|---|---|---|---|---|
| Light Sensors | W | 26.9 | **24.8** | **24.2** | **23.1** | **24.7** |
| | F | **26.8** | 28.3 | 26.7 | 26.8 | 27.1 |
| | A | 33.8 | 29.2 | 28.9 | 30.0 | 30.5 |
| | C | 27.7 | 25.4 | 24.4 | 25.1 | 25.7 |
| Temperature Sensors | W | 27.9 | **23.2** | 26.0 | **23.1** | 25.1 |
| | F | 28.6 | 26.6 | 27.3 | 27.0 | 27.4 |
| | A | 31.9 | 28.6 | 30.9 | 28.8 | 30.1 |
| | C | **23.7** | 24.5 | **23.8** | 25.6 | **24.4** |

Table 5

RMSE (%) of predictions for continuous sensors using Wavelet (W), FreMEn (F), ARIMA (A) and CNN-based (C) models in the ENRICHME dataset

| | | Bathroom | Bedroom | Kitchen | Living Room | Average |
|---|---|---|---|---|---|---|
| Light Sensors | W | **20.4** | 21.9 | **20.5** | **21.0** | **20.9** |
| | F | 22.2 | 21.6 | 21.7 | 21.9 | 21.9 |
| | A | 31.6 | 30.3 | 30.8 | 29.1 | 30.4 |
| | C | 22.4 | **20.6** | 21.5 | **21.0** | 21.4 |
| Temperature Sensors | W | 22.3 | 21.0 | 22.0 | **20.2** | **21.4** |
| | F | **21.6** | 22.7 | 23.1 | 22.0 | 22.3 |
| | A | 34.7 | 29.8 | 29.8 | 30.2 | 31.1 |
| | C | 23.7 | **20.9** | **20.8** | 20.5 | 21.5 |

based upon a fully-convolutional neural network (CNN) [3] using causal convolutions without feedback. This takes a sequence of any length and map it to an output sequence of the same length.

To start with, Table 2 presents some statistics of the predictions in the L-CAS dataset. For all the considered metrics, we can see that our new wavelet model outperforms the others. In particular, the wavelet model performs much better in terms of accuracy. Table 3 presents also some results on the ENRICHME dataset. In this case, the precision of the FreMEn and CNN models are slightly higher than our wavelet model, probably due to the periodic nature of the activities in the considered scenario. FreMEn indeed captures all the most relevant frequency components, so the predicted activations can be very precise (i.e. high number of true positives). However, for the recall, which considers the correct predictions over the total number of real activations, we can observe a significant improvement of the wavelet models compared to CNN-based and FreMEn models, since the latters are not able to predict some of the sensor activations. This improvement is further confirmed by the F1 score and the accuracy, also shown in the same table.

Continuous signals such as those from light and temperature sensors are also properly modelled by our wavelet models. The average RMSE of the prediction is the lowest for the light sensors in the L-CAS dataset (see Table 4). Only in one occasion, i.e. for the entry sensor, the value predicted by the FreMEn model is slightly better than ours, most likely due to its significant periodic component. The performance of wavelet models is particularly relevant in this case. The accuracy of the model is key for our anomaly detection system, as it will define how different rules are triggered. The temperature predictions are very similar for the wavelet and the CNN-based models, although the latter is more accurate for the entry and the lounge sensors, probably thanks to the large amount of training data available in this dataset. The ARIMA models perform reasonably well but, as expected, not as good the other two.

The results for the ENRICHME dataset (see Table 5) show a similar situation. The wavelet models are generally better than the FReMEn and the CNN-based ones, although the difference to the latter is minimal. Note also that all the considered models perform better in this one than the L-CAS dataset. In fact, the ENRICHME dataset captures the interactions of only two persons in a small domestic environment, while the L-CAS dataset was recorded in a larger and more crowded office environment.

### *6.3. Performance of activity representation*

In the following sub-sections we illustrate the performance of our system to represent human activities using the normalized entropy method from [22] and comparing the expected levels of activity to the actual ones.

#### *6.3.1. Real vs. predicted entropies*

We compared the entropies of human activity predicted by our wavelet model with the actual ones computed on both datasets. We used three popular metrics to measure the statistical similarity between these two entropies: RMSE, correlation coefficient, and explained variance.

Table 6 illustrates the good performance of our solution in predicting the entropy of human activities, showing better results than the other approaches. We can also see that the entropy predicted by our wavelet model is mostly improved for the ENRICHME dataset compared to the L-CAS dataset (i.e. lower RMSE; higher correlation and explained variance). Only the RMSE is slighly better for the CNN-based models in the ENRICHME dataset. However, for both datasets, our results confirms that real and predicted entropies are reasonably similar and, therefore, the wavelet-based model is suitable to forecast the level of activity in the environment.

Table 6

Measures of similarity between real and predicted entropies of human activity in the L-CAS and ENRICHME datasets using Wavelet (W), FreMEn (F), ARIMA (A) and CNN-based (C) models

| Dataset | Model | RMSE (%) | Correlation (%) | Explained Variance (%) |
|---|---|---|---|---|
| L-CAS | W | **23.1** | **68.0** | **36.2** |
| | F | 25.7 | 60.5 | 16.2 |
| | A | 28.6 | 54.7 | 14.7 |
| | C | 23.7 | 67.2 | 27.1 |
| ENRICHME | W | 20.2 | **74.2** | **51.6** |
| | F | 21.1 | 65.2 | 27.6 |
| | A | 28.0 | 68.3 | 36.0 |
| | C | **20.0** | 71.5 | 40.3 |

#### *6.3.2. Examples of activity forecasting*

As explained in [ 22], human activities can be represented by the normalized entropy of the environment. Figure 5 illustrates two examples of such entropy calculated from the real sensors and predicted by our wavelet-based model. In particular, the red graph shows the real normalized house entropy (as percentage) based on the available sensor setups. The blue graph is the predicted entropy at the same time, using the wavelet models of our sensors. It also shows the entropy prediction based on FreMEn (green), ARIMA (yellow) and CNN-based (pink) models. We can see the limitations of FreMEn, which is not capable of capturing the complex frequency behaviours of the abrupt slope in Fig. 5a or the low/high combination in Fig. 5b. The ARIMA model suffers from similar frequency limitations, while the CNN-based model shows a better performance, closer to our wavelet solution.

More in detail, Fig. 5a is based on the ENRICHME dataset, collected in the relatively quiet apartment of an elderly couple. The figure refers to a typical morning of the two residents. The predicted entropy of their activities differs from the real one for less than 10%, with only two significant exceptions: in the morning, at around 10:00, the activity's level was higher than expected (about 20% error between real and predicted entropies); a little later, around 11:30, the real activity's entropy decreased sharply a few minutes after the usual time (still about 20% error). These differences between real and predicted data, however, are understandable under normal variations of the resident's schedule, which cannot be predicted by our model. It is worth noticing that our solution is able to predict a very sharp transition, where the activity's entropy goes from high to no activity at all. Other models are limited in their responsiveness, showing the improved capability of our system to consider high-frequency elements thanks to its wavelet-based model.

Figure 5b refers instead to the activity of a non-typical Friday afternoon in the L-CAS offices. The real entropy (red) shows that it was a particularly busy day, with a high activity level for most of the time. However, a significant decrease of the entropy between 18:00 and 19:00, when most of the researchers left the office, is followed by another increase between 19:00 and 20:00, when some people came back. The activity remained then relatively high for the rest of the evening, which was unusual. The entropy's prediction (blue) is able to capture several important trends of the activity levels, including a few small negative peaks between 17:00 and 18:00 hours, probably due to some researchers leaving the office, and the sharp decrease around 18:00 hours, when most of them left. Our model captures also some of the evening activities and the entropy's increase between 19:00 and 20:00. Although after this time there is a significant difference between real and predicted entropies, due to the unusual presence of people on a Friday night, the general trends of the activity's levels are correctly captured by our prediction system.

### *6.4. Performance of anomaly detection*

In this final set of experiments we compare our HMLN for anomaly detection (Section 4 and 5), which integrates wavelet and entropy-based activity priors, to other existing approaches. Abnormal activity detection ensures timely intervention and helps mitigate more significant problems, as elderly health incidents and food damage. To achieve the latter, some emergency assistance methods were introduced when sensors readings are classified as abnormal. Common emergency methods include among others alarm systems, email notifications, SMS notifications. In this work, anomalies are reported using ROS services both to the assistive robot [14] and networked care [4]. This means a double approach is taken for emergency assistance: an inmediate robot interaction is requested while common emergency methods are addressed by the networked care system.

The normalized entropy computed by our system can be used indeed as a time series for unsupervised anomaly detection. In particular, we evaluate our HMLN-based anomaly detector against five state-of-the-art unsupervised methods from existing frameworks [6,19,33,59,60], which are the following ones:

- Gaussian1D [10] – A frequentist anomaly detection method that assumes the intput data is Gaussian, searching for low likelihood values [60].
- LOF [10] – This method relies on local deviations of the density of a given sample with respect to its neighbors. It is local in the sense that the anomaly score depends on how isolated the object is from the surrounding neighborhood [19].

[10] https://github.com/MentatInnovations/datastream.io

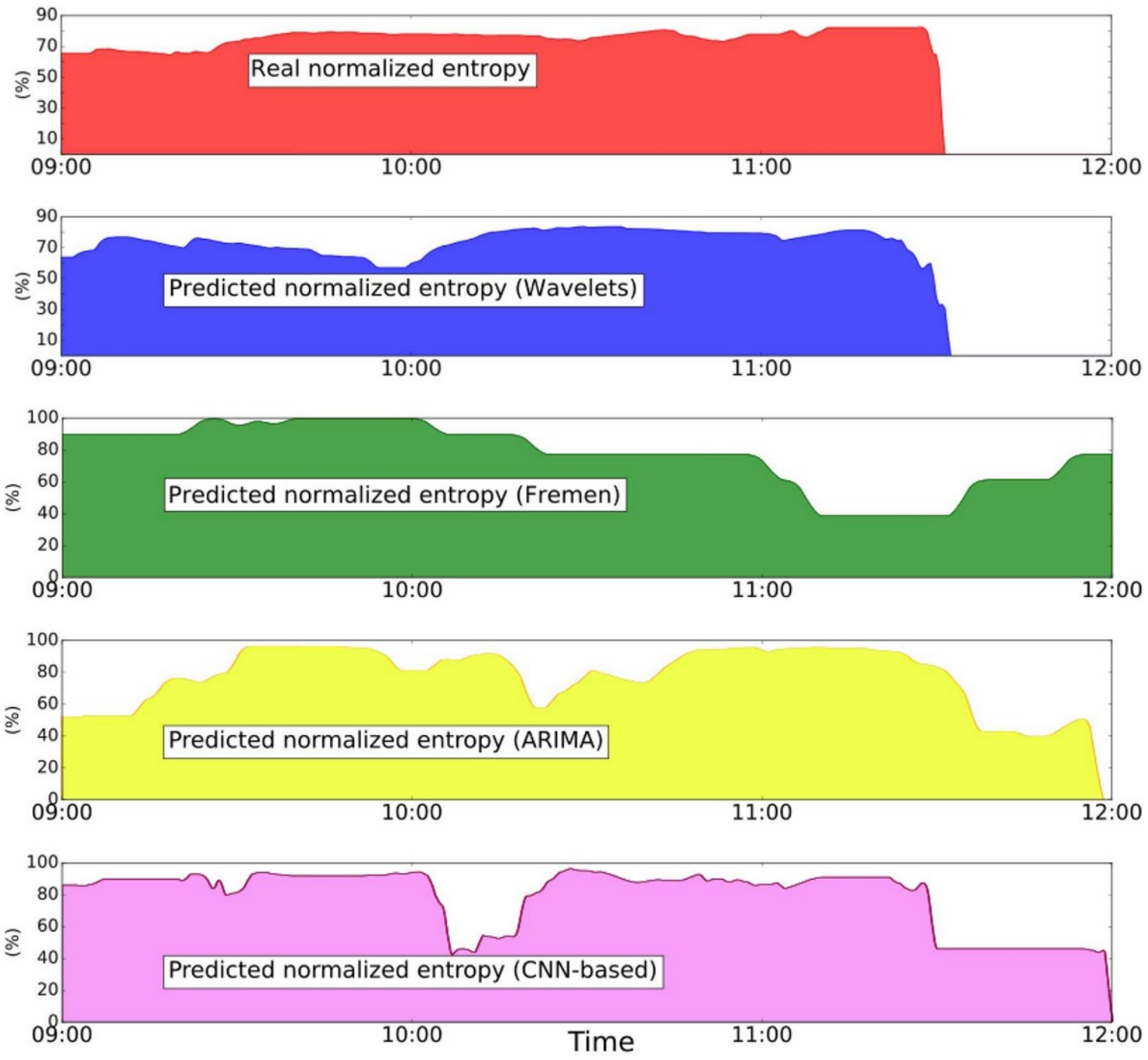

(a)

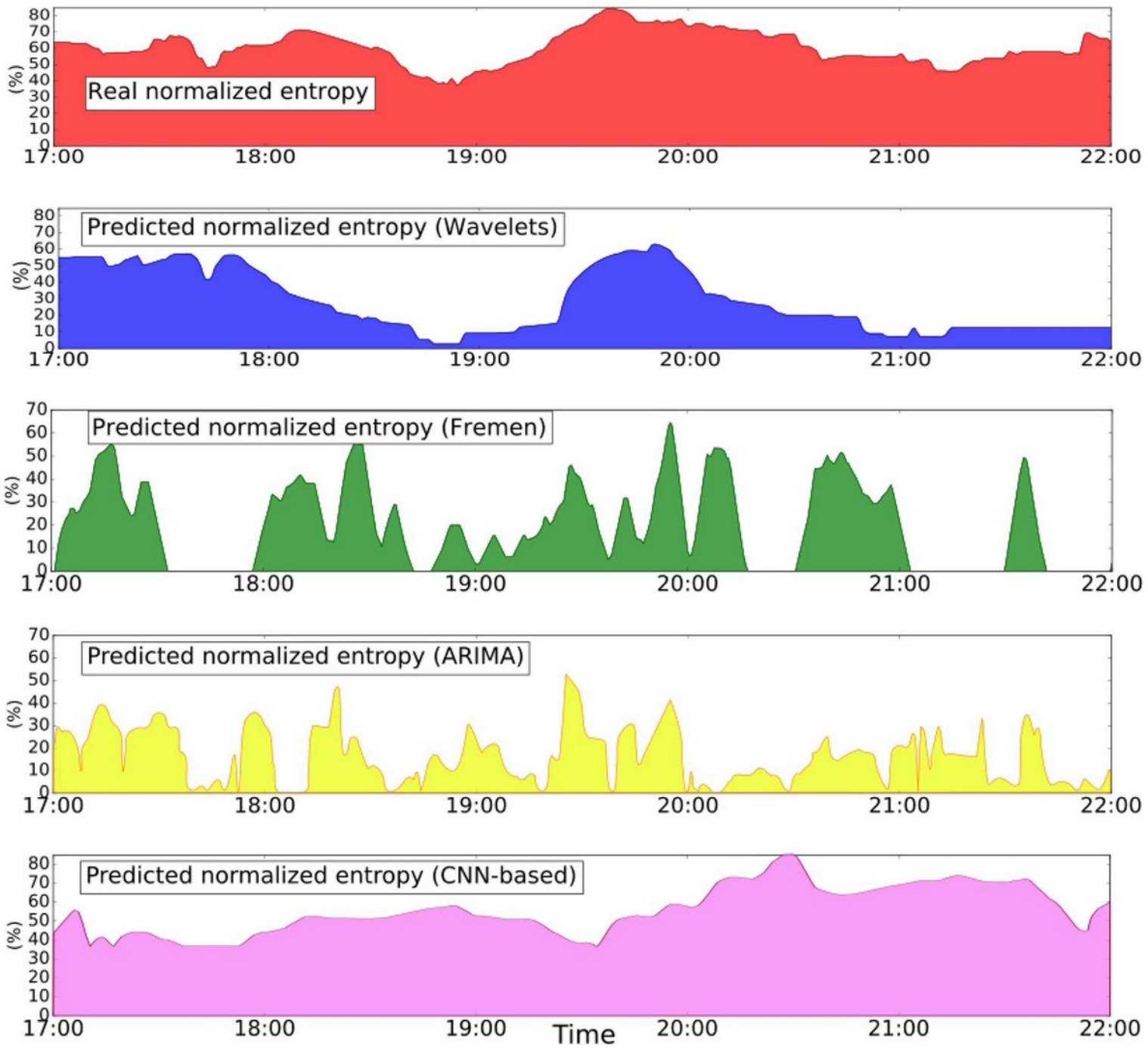

(b)

Fig. 5. Examples of real vs. predicted entropies computed on the ENRICHME (a) and L-CAS (b) datasets during a 5-hours interval. The first plot's row, in red, depicts real entropies. The second row, in blue, is the wavelet-based prediction. The third row, in green, shows the FreMEn forecast. The forth row, in yellow, shows the ARIMA estimation and last row, in pink, shows a CNN-based model output.

Table 7

Percentage of anomalies detected by a particular method (row) that are also detected by another one (column) in the L-CAS dataset

| Anomalies in | Also detected by: Gaussian1D | LOF | COPOD | LSCP | $\beta$-VAE | HMLN | HMLN* |
|---|---|---|---|---|---|---|---|
| Gaussian1D | 100 | 14.4 | 32.3 | 11.6 | 21.3 | 10.2 | 9.8 |
| LOF | 3.0 | 100 | 4.8 | 21.5 | 11.1 | 21.5 | 21.0 |
| COPOD | 25.8 | 9.3 | 100 | 18.3 | 100.5 | 8.7 | 8.0 |
| LSCP | 7.7 | 17.7 | 10.5 | 100 | 17.0 | 19.2 | 19.3 |
| $\beta$-VAE | 17.4 | 5.7 | 61.4 | 17.2 | 100 | 9.1 | 8.6 |
| HMLN | 3.4 | 31.3 | 7.1 | 23.3 | 9.8 | 100 | 99.5 |
| HMLN* | 2.0 | 31.3 | 7.8 | 24.0 | 9.5 | 100 | 100 |

Table 8

Percentage of anomalies detected by a particular method (row) that are also detected by another one (column) in the ENRICHME dataset

| Anomalies in | Also detected by: Gaussian1D | LOF | COPOD | LSCP | $\beta$-VAE | HMLN | HMLN* |
|---|---|---|---|---|---|---|---|
| Gaussian1D | 100 | 25.1 | 40.3 | 52.0 | 31.7 | 77.9 | 65.8 |
| LOF | 14.9 | 100 | 26.8 | 53.7 | 10.8 | 21.4 | 18.2 |
| COPOD | 36.8 | 36.3 | 100 | 47.8 | 46.3 | 80.6 | 69.5 |
| LSCP | 10.7 | 38.7 | 26.7 | 100 | 13.5 | 34.5 | 29.4 |
| $\beta$-VAE | 11.4 | 4.6 | 15.2 | 8.6 | 100 | 32.9 | 30.0 |
| HMLN | 38.9 | 19.1 | 52.6 | 36.7 | 43.6 | 100 | 84.5 |
| HMLN* | 39.2 | 18.4 | 52.4 | 34.7 | 43.1 | 100 | 100 |

– COPOD [11] – A parameter-free, unsupervised outlier detection algorithm inspired by statistical methods for modeling multivariate data distributions [33].
– LSCP [11] – Another unsupervised method that defines local regions around test instances using the consensus of its nearest neighbors in randomly selected feature subspaces. The top-performing base detectors in these local regions are selected and combined as the model's final output for detection [59].
– β-VAE[11] – A variational autoencoder which uses a modification to the training regime to improve the learning of disentangled representations [6].

To compare our HLMN to the above methods, we account for the number of anomalies that each detector has in common with the others. The results are summarized in Table 7 and 8 for the L-CAS and ENRICHME datasets, respectively. For a fair comparison, the tables include also a variant of our method (HMLN*) that does not implement any expert rule, but considers only statistical anomalies based on activity entropy. We can see that all the anomalies reported by the HMLN* with no rules are also reported by the original HMLN, but not the opposite, as expected.

The results show that our HMLN approach shares a significant number of detections with the other five methods. In particular, our solutions enable a balanced distribution of detections that captures a reasonable number of anomalies from the other detectors. It is also interesting to note how the HMLN detectors behave similarly to the LOF and the LCSP detectors for the L-CAS dataset (Table 7), whereas for the ENRICHME dataset (Table 8) the HLMN detectors are closer to COPOD and β-VAE. This difference may come from the fact that each dataset captures a very different scenario: the first one is a research office, while the second is an elderly home. The type of anomalies on each dataset is different, an thus they are detected by different algorithms, although the HMLN detectors are still able to perform reasonably well in both cases.

To identify the best one among these detection systems, but lacking a consistent and reliable annotation of true anomalies, we used the method proposed by Lamiroy & Sun [31] to estimate precision and recall, and from these compute the F1 score. Although not accurate in absolute terms, this approach has been shown to be useful for ranking different binary classifiers in absence of ground-truth.

[11] https://github.com/yzhao062/pyod

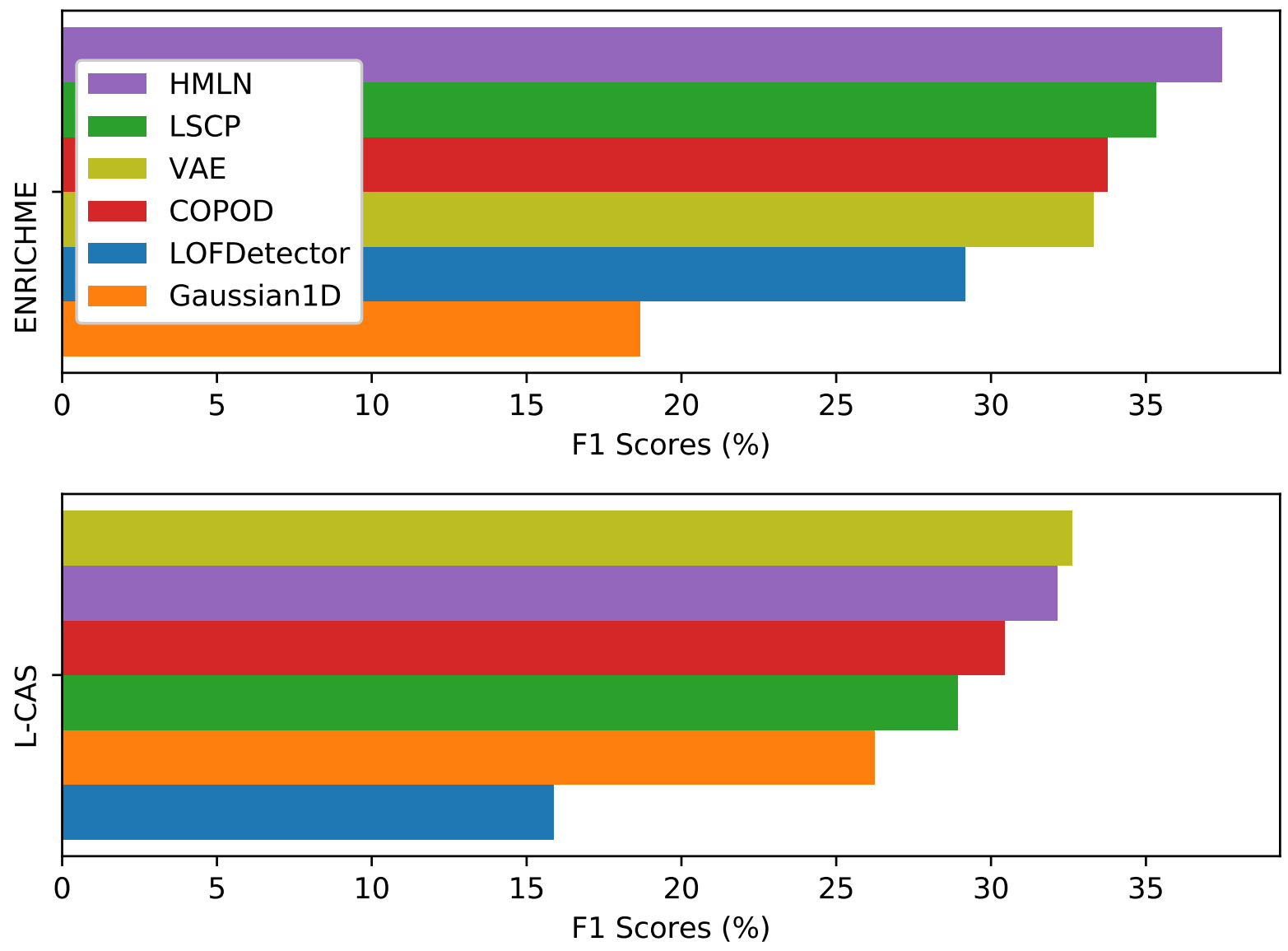

(a)

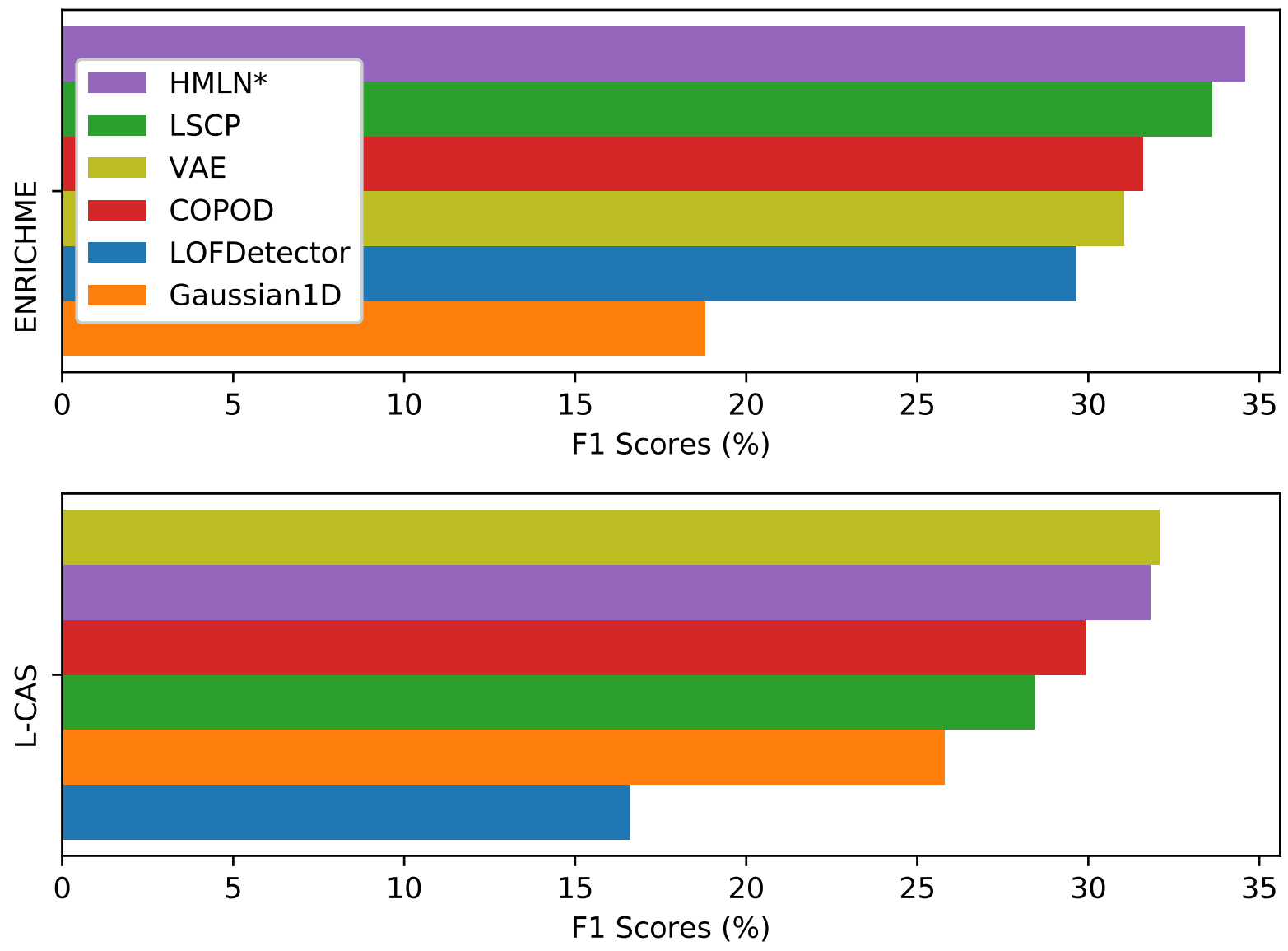

(b)

Fig. 6. F1 score comparison for the considered methods using our HMLN anomaly detector with (a) and without (b) expert rules.

Lamirov's method was applied to our datasets to obtain the results summarised in Fig. 6. In particular, we used F1 estimations to rank different detectors in absence of ground truth. Note that the ranking is the same in Fig. 6a (with expert rules) and Fig. 6b (without), although the F1 scores obviously differ in terms of numerical values.

Also, although in general our HMLN detectors are the most performant, we should mention that the β-VAE encoder ranks slightly above them in the L-CAS dataset. This could be due to the challenging nature of the dataset itself, which contains many more people, sensors, and time entries. As already seen in previous results, in this case it is more difficult to identify periodic patterns of sensor data, and hence anomalies. Therefore, a neural network-based detector might be more suitable to model such complex scenarios, provided sufficient training data (3 months in the L-CAS case) is available.

## 7. Conclusions and future work

This paper presented a new approach for wavelet-based temporal modelling of smart sensors, which we used to forecast levels of human activity in dynamic and robot-assisted environments. We also proposed an original application of HMLNs combining real and predicted entropies of human activity with expert rules to detect potential anomalies. Our solutions have been evaluated using two large public datasets, one of which newly collected from a real elderly home, to demonstrate their effectiveness.

We have compared our wavelet-based activity models and anomaly detection to several state-of-the-art solutions, showing improved overall performance. In particular, the wavelet model performed better than other three popular forecasting approaches, while the anomaly detection system has proved to be preferable when compared to alternative unsupervised methods. It is also interesting to note that our solution can be a valid alternative to neural network-based approaches in scenarios with limited amount of data, like the ones here considered.

Although the proposed wavelet temporal model can be applied to any arbitrary signal, our current implementation focused mostly on static binary sensors, partly because of the subsequent entropy-based representation of human activities. It remains to be studied how analog smart sensors (e.g. light, temperature) and mobile sensors (e.g. from the robot) could be integrated and exploited by our system.

Finally, despite the flexibility of HMLNs, there are still limitations in the way logic rules are formulated and their weights learned, which requires particular attention and fine tuning to guarantee the convergence of the training process. Also, the time required by the latter grows exponentially with the number and complexity of the rules, which can be a problem in case a richer spectrum of human activities and sensor data is considered. In our case, indeed, the MLN's rules are relatively simple to guarantee the real-time performance of our system, an aspect that we analysed in previous work [21]. Possible alternatives combining deep neural networks and symbolic representations, like Logic Tensor Networks [44] and ABBA-LSTM [17], could potentially overcome some of these problems and enable more powerful inference systems for anomaly detection. In addition to the above-mentioned integration of analog and mobile sensors, future work will be therefore devoted to more recent and promising neuro-symbolic architectures [36], with special attention to the constraints posed by data availability and real-time requirements.

## Conflict of interest

The authors have no conflict of interest to report.